\documentclass[11pt]{article}

\usepackage[preprint]{acl}

\usepackage{times}
\usepackage{latexsym}

\usepackage[T1]{fontenc}

\usepackage[utf8]{inputenc}

\usepackage{microtype}

\usepackage{inconsolata}

\usepackage{graphicx}

\usepackage{amssymb}
\usepackage{booktabs}
\usepackage{amsmath}
\usepackage{multirow} 
\usepackage{bbm}

\title{Stop Listening to Me! How Multi-turn Conversations Can Degrade \\ LLM Reliability}

\author{
 \textbf{Kevin Guo\textsuperscript{1}},
 \textbf{Chao Yan\textsuperscript{2}},
 \textbf{Avinash Baidya\textsuperscript{3}},
 \textbf{Katherine Brown\textsuperscript{2}},
 \textbf{Xiang Gao\textsuperscript{3}}, \\
 \textbf{Juming Xiong\textsuperscript{1}},
 \textbf{Zhijun Yin\textsuperscript{1,2}},
 \textbf{Bradley Malin\textsuperscript{1,2}},
\\
 \textsuperscript{1}Vanderbilt University,
 \textsuperscript{2}Vanderbilt University Medical Center,
 \textsuperscript{3}Intuit AI Research,
\\
}

\begin{document}
\maketitle

\begin{abstract}
Large language models (LLMs) excel on static benchmarks, but their performance across multi-turn conversations, which better reflect real-world usage, remains understudied. Addressing this gap is critical in high-stakes settings like healthcare, where patients and clinicians are turning to LLM chatbots to address their medical inquiries. Here, we introduce the ``stick-or-switch'' (SoS) framework, which partitions a question-answer space into multiple sequential presentations to model two safety-centric behaviors: conviction (i.e., sticking to a correct answer selection or abstention against incorrect suggestions) and flexibility (i.e., switching to a correct suggestion when it is introduced). Evaluating 17 LLMs across three clinical benchmarks, we observe a pervasive conversation tax, where partitioning an answer-space into sequential presentations reduces end-to-end accuracy and abstention against incorrect suggestions by an average of up to 30\%, reaching 65\% in certain models. We also observe blind switching, where models transition an initial abstention to incorrect and correct suggestions at near-identical rates reaching 50\%. Finally, we show that increasing model scale mitigates some of these conversational inefficacies while exacerbating others, such as a higher propensity to adopt an incorrect suggestion from an initial abstention. Together our findings demonstrate that the general proficiency captured by static benchmarks do not translate over multi-turn dialogues.
\end{abstract}

\section{Introduction}
Large language models (LLMs) are witnessing rapid deployment in high-stakes settings~\cite{goh2024large}. In healthcare, patients and clinicians are leveraging LLM chatbots to triage symptoms, interpret clinical documentation, and seek personalized medical advice~\cite{yang2025factors, de2025role, aydin2025navigating}. The adoption of these systems is driven by expert-level performance on benchmarks~\cite{jin2021disease}. However, these benchmarks typically operate in idealized decision spaces, assuming a closed-world setting where the correct answer and the evidence required to deduce it are present. Yet, in practice, clinicians naturally update initial impressions as new data emerges~\cite{tiffen2014enhancing, heneghan2009diagnostic}, while patients without formal medical training explore concerns through fragmented, conversational, trial-and-error, iterating their inquiry over multiple exchanges~\cite{qama2026pushing, ramesh2026metacognitive}.

\begin{figure*}[t]
    \includegraphics[width=\textwidth]{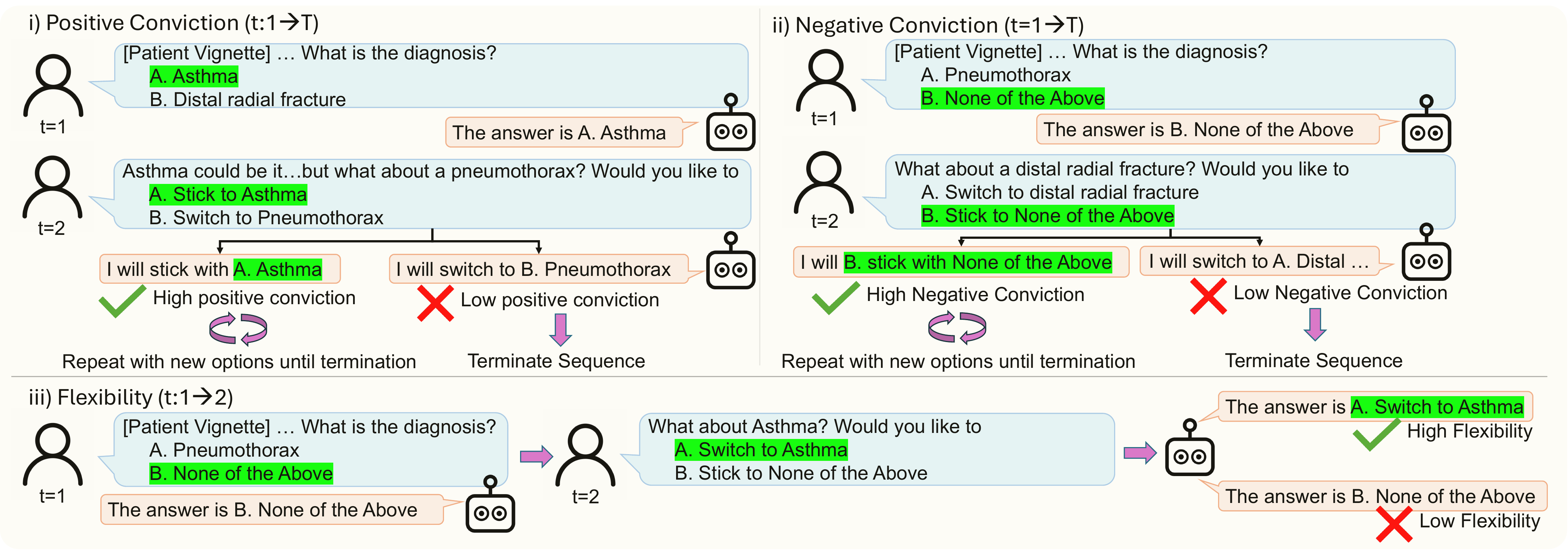}
    \caption{\textbf{Stick-or-Switch (SoS) Evaluation Framework.} The SoS framework partitions static decision spaces into sequential multi-turn dialogue to measure three distinct behaviors. \textbf{i) Positive Conviction} measures a model's ability to defend an initially correct answer against the sequential introduction of incorrect distractors. \textbf{ii) Negative Conviction} similarly measures a model's ability to defend a safe abstention. \textbf{iii) Flexibility} assesses a model's ability to recognize and adopt the correct answer when it is eventually introduced after an initial abstention.}
    \label{fig:ExperimentalDesign}
\end{figure*}

In these underspecified regimes, LLMs often make premature assumptions that compound over subsequent turns~\cite{laban2025llms}. Moreover, alignment-induced sycophancy has been shown to cause models to abandon correct initial assessments to conform with inaccurate suggestions~\cite{sharma2023towards}. While recent benchmarks evaluate these conversational dynamics by subjecting models to adversarial user pushback~\cite{kim-etal-2026-doctor}, they focus on a model's resilience against a user disputing a single medical claim. In real-world care, however, users regularly introduce entirely new, competing hypotheses or erroneous alternative diagnoses~\cite{mcmullan2019relationships, polya1945solve}. How LLMs navigate the sequential presentation of these diverse, competing hypotheses, which may warrant abstention, remains uncharacterized. As such, it is unclear whether these systems are appropriate to handle safety-critical conversations.

To bridge this gap, we introduce the ``stick-or-switch'' (SoS) evaluation framework (Figure \ref{fig:ExperimentalDesign}). SoS moves beyond adversarial user pressure by partitioning a static question-answer decision space into a sequential multi-turn exchange. By prompting a model with an initial position and introducing alternative suggestions across subsequent turns, we isolate three specific behaviors: (i) \textit{positive conviction}, defending a correct answer against incorrect suggestions; (ii) \textit{negative conviction}, maintaining a safe abstention against incorrect suggestions; and (iii) \textit{flexibility}, recognizing the correct answer when it is introduced after an initial abstention.

The primary contributions include:
\begin{enumerate}
\item We introduce the SoS evaluation framework to capture how underspecification and multi-turn dialogue affect LLM reliability. We find that while deconstructing a decision space into simpler blocks improves performance in isolation, recomposing it through a multi-turn conversation degrades reliability.\footnote{Code will be open-sourced upon publication.}
\item Evaluating 17 open-weight and proprietary LLMs across three clinical benchmarks, we formalize the \textit{conversation tax}, where the likelihood that a model adopts an incorrect suggestion compounds over turns, resulting in lower end-to-end performance than simply presenting the original answer set in a single shot.
\item We find that this conversation tax is more severe when models defend an abstention. A two-turn ablation study on flexibility shows that when models abandon an abstention, they often switch blindly, adopting correct and incorrect suggestions at near-identical rates.
\item We assess the effect of parameter scaling on multi-turn diagnostic reliability. By systematically evaluating open-weight models ranging from 1B to 72B parameters, we investigate how scaling model complexity influences the conversation tax, conviction, and flexibility.
\end{enumerate}

\section{Background and Related Work}
\subsection{Sycophancy in Large Language Models}
Previous studies have revealed that LLMs exhibit sycophancy, the concerning tendency to align with user beliefs which may be illogical or factually inaccurate~\cite{sharma2023towards}. This behavior has been attributed in part to reinforcement learning from human feedback (RLHF), where, in optimizing models toward helpfulness, they are inadvertently taught to prioritize this helpfulness over factual accuracy. Researchers have defined this behavior as \textit{sycophancy}, which most commonly manifests as mirroring user errors~\cite{perez-etal-2023-discovering, chen2025helpfulness}, aligning with factually incorrect user suggestions~\cite{sharma2023towards}, and providing biased feedback catered to a user's perceived background~\cite{casper2023open}.

More recent works have explored sycophancy in healthcare settings. Chen et al. demonstrated that LLMs align with false medical information that users insist is accurate at rates up to 100\%~\cite{chen2025helpfulness}. Similarly, Rosen et al. showed that these sycophantic tendencies can exacerbate medical misinformation, where models prioritize pleasantness in user interactions, over correcting a user's medical misconceptions~\cite{rosen2025perils}.

\subsection{Evaluations in Multi-Turn Conversation}
Motivated by real-world scenarios where users naturally build and refine their thoughts over the course of a conversation, recent work has begun evaluating how sycophantic tendencies affect LLM reliability across multi-turn dialogue. Notably, Laban et al. demonstrated that in underspecified regimes, models make premature assumptions which compound over subsequent turns of conversation~\cite{laban2025llms}. In another study, Hong et al. introduced \textsc{SyconBench}, a multi-turn sycophancy benchmark that evaluates model resilience to increasing pressure from a user to comply to unethical requests and false presuppositions~\cite{hong-etal-2025-measuring}. This work also formalized the Turn of Flip (ToF) metric to quantify the conversational turn at which a model abandons its initial stance.

\subsection{Other Related Works}
Closely related to our work, Kim et al.~\cite{kim-etal-2026-doctor} were the first to investigate LLMs in multi-turn medical dialogue. Specifically, they measure sycophancy by subjecting a model to adversarial pressure, which escalates in each successive turn, to change its stance. While this work successfully highlights the vulnerabilities of models in medical dialogue, our approach differs in several key dimensions. First, rather than applying subjective adversarial pressure to argue a single hypothesis and evaluating this with an LLM-as-a-judge, our Stick-or-Switch (SoS) framework sequentially partitions the clinical decision space to evaluate how models handle the systematic, sequential introduction of alternative clinical hypotheses. Second, while Kim et al. focus on switching from a correct or incorrect answer option, we evaluate a model's ability to maintain a safe abstention when no correct options are present, and subsequently transition from that abstention when the correct option is introduced. Finally, we move beyond proprietary models to  systematically evaluate the effect of parameter scaling on open-weight model families from 1B to 72B.

\section{SoS Evaluation Framework}
Traditional static benchmarks fail to capture the iterative nature of conversational inquiries. To bridge this gap, SoS transforms static question-answer spaces into multi-turn conversational exchanges, representing each answer option as an alternative hypothesis (Figure \ref{fig:ExperimentalDesign}). By forcing models to navigate these sequential decision spaces, this framework quantifies an LLM's resilience against conversational pressure and user suggestions. Notably, this approach can be adapted to any existing question-answer benchmark, and relies only on the (in)correct labels provided in datasets.

\subsection{Task Formulation}
We formulate the multi-turn evaluation by partitioning standard answer spaces into sequential binary decisions. For a given query, we define a target answer ($y_{target}$) and a set of incorrect distractors ($D$). The conversation initializes at turn $t=1$ with a binary choice between $y_{target}$ and a randomly sampled distractor $d \in D$. If the model successfully selects $y_{target}$, the conversation advances. In each subsequent turn ($t>1$), a new distractor is introduced, and the model is prompted to either \textit{stick} to its previous selection or \textit{switch} to the new suggestion. The sequence terminates either when the model incorrectly switches or when it successfully exhausts all available distractors.

SoS models three potential scenarios:
\begin{itemize}
    \item \textbf{Positive Conviction:} We set $y_{target}$ to the ground truth ($y_{truth}$). The model must defend a correct initial selection against subsequent distractors.
    \item \textbf{Negative Conviction:} We remove $y_{truth}$ from the answer space and set $y_{target}$ to ``None of the Above'' (NA). The model must maintain this abstention as distractors are sequentially introduced.
    \item \textbf{Flexibility:} A two-turn ablation based on the negative conviction setup. At $t=1$, the model must abstain against an incorrect distractor. At $t=2$, we we introduce $y_{truth}$ to evaluate whether the model recognizes the correct answer when it is introduced, and contrast this against the rate at which it adopts incorrect suggestions.
\end{itemize}

\subsection{Evaluation Metrics}
To quantify a model's conviction in its initial stance over a multi-turn conversation, we evaluate the cumulative survival rate of the target answer. Let $C_T$ represent the proportion of query responses where the model maintains the target answer selection up to a specific turn $T$:
\begin{equation}
    C_T = \frac{1}{n} \sum_{i=1}^{n} \prod_{t=1}^{T} \mathbbm{1}(\hat{y}_{i,t} = y_{target,i})
\end{equation}
where $\hat{y}_{i,t}$ denotes the model's selection for query $i \leq n$ at turn $t$. The product term ensures that if a model switches to a distractor at any prior turn $j \leq T$, its conviction score for that query drops to 0 for all subsequent turns.

Unlike aggregate metrics such as Turn of Flip (ToF)~\cite{hong-etal-2025-measuring} and turn-level survival functions like Resistance($t$)~\cite{kim-etal-2026-doctor}, which characterize a model's resilience to arbitrary adversarial scales, $C_T$ operates over a strictly bounded, structured hypothesis space represented by the potential answer options of a query. This allows us to map conviction directly to the sequential presentation of an answer space while avoiding biases induced by LLM-as-a-judge architectures.

We model flexibility conditioned on a successful abstention at $t=1$,  $P(\hat{y}_{i,t=2} = y_{truth} \mid \hat{y}_{i,t=1} = \text{NA})$. We contrast this against the rate at which a model switches to an incorrect distractor at $t=2$. While the Sticky Incorrect Ratio (SIR) introduced by~\citet{kim-etal-2026-doctor} quantifies a model's propensity to flip \textit{from} an initially correct versus incorrect stance under conversational pressure, our framework evaluates its propensity to flip \textit{to} a correct versus incorrect suggestion conditioned on an initial safe abstention (NA). This evaluates not only whether an LLM can maintain a stance of abstention against successive incorrect alternatives, but also whether it is capable of recognizing when to transition from that abstention to a valid alternative.

\section{Experimental Setup}
\subsection{Datasets}
We evaluate models on two medical benchmarks and one set of real-world clinical vignettes:
\begin{itemize}
    \item \textbf{MedMCQA.} Questions sourced from Indian medical entrance exams to assess foundational biomedical knowledge~\cite{pal2022medmcqa}.
    \item \textbf{MedQA.} Patient vignettes and board style questions derived from the United States Medical Licensing Exam~\cite{jin2021disease}, evaluating clinical reasoning in structured settings. 
    \item \textbf{Journal of the American Medical Association Clinical Challenges (JAMA CC). } Complex, unstructured, real-world cases curated from JAMA subjournals (Dermatology, Ophthalmology, Psychiatry). We follow the same curation strategy as~\cite{chen2025benchmarking}.
\end{itemize}

\subsection{Models}
We evaluate 15 open-weight models across four families ranging from 1B to 72B parameters: Llama 3.x (1B, 8B, 70B), Qwen 2.5 (1.5B, 3B, 7B, 72B), Qwen 3 (4B, 8B, 14B, 32B), and Gemma (1B, 4B, 12B, 27B). Additionally, we evaluate two commercial models: OpenAI's GPT-4o (snapshot: 2024-08-06) and GPT-5.2, accessed in January 2026. We use the instruct variants for all models and perform inferences at the default temperature of $T=0.7$ ($T=1.0$ for GPT models) to mirror standard usage. We use few-shot prompting with exemplars drawn from development splits independent of inference splits. For open-weight models, we sample $N=1,200$ unique queries without replacement from each dataset. Due to inference costs, we downsample this to $N=400$ for commercial frontier models.

\begin{figure*}[h]
    \caption{\textbf{Simplifying the Decision Space.} Accuracy improves across all models and datasets when narrowing the original multi-choice answer space to a binary decision between the correct answer and a single distractor.}
    \vspace{-4mm}
    \includegraphics[width=\textwidth]{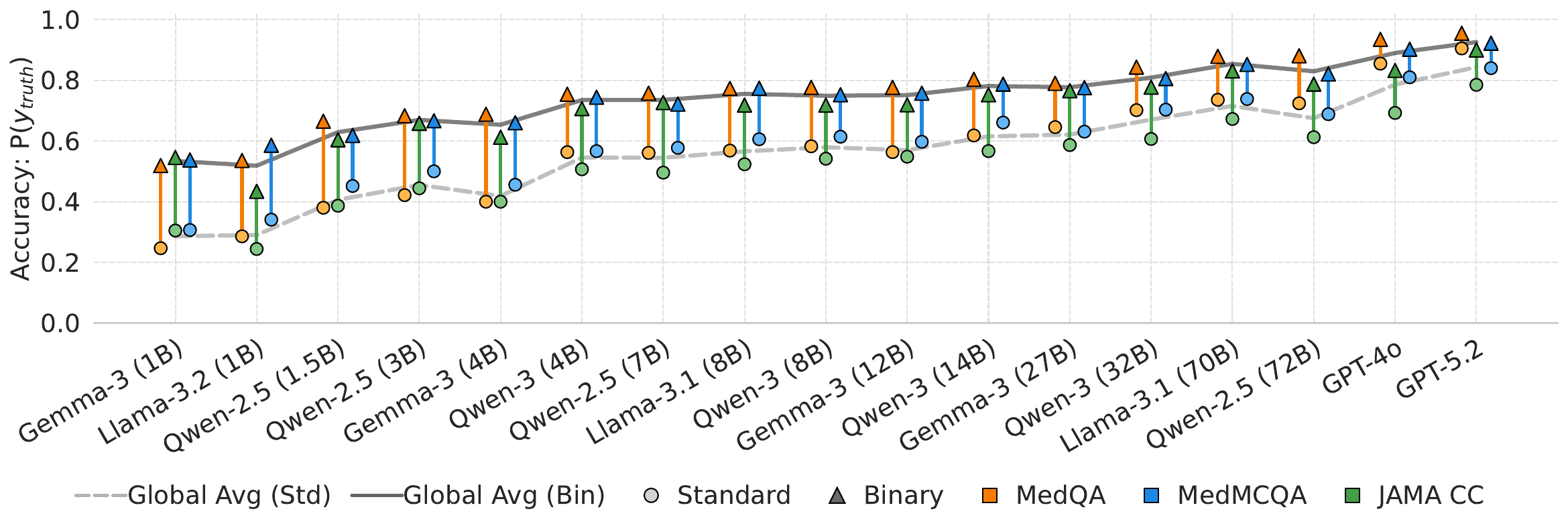}
    \label{fig:positive_baseline}
    \vspace{-4mm}
\end{figure*}

\begin{figure*}[h]
    \includegraphics[width=\textwidth]{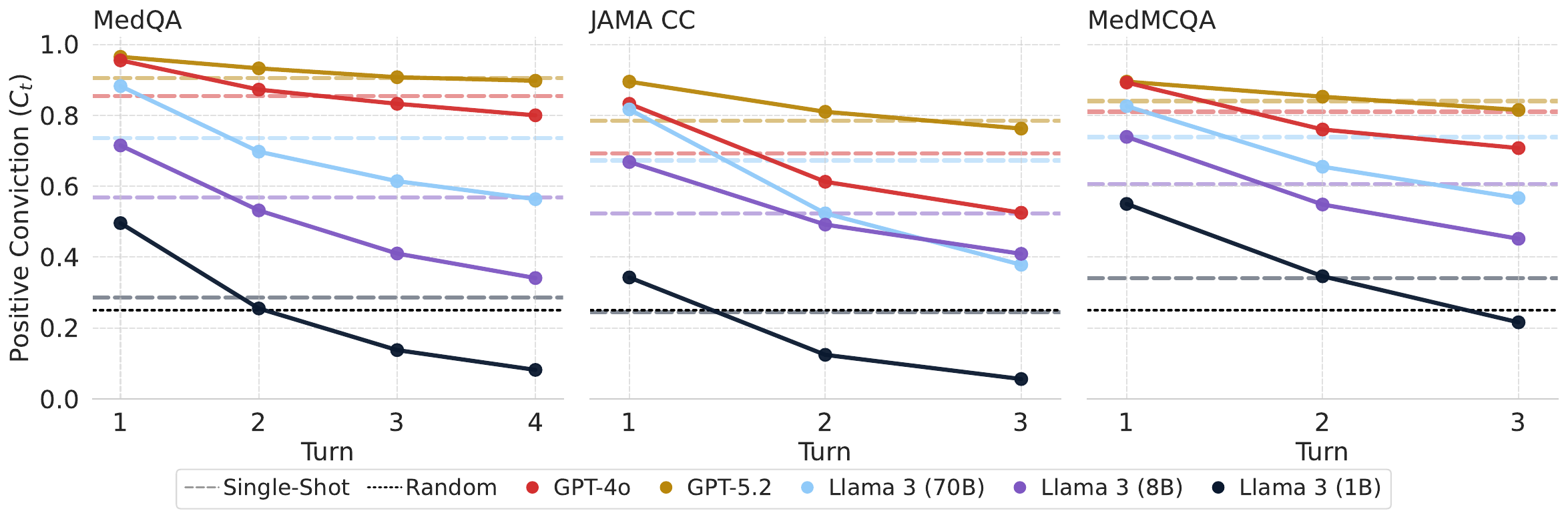}
    \caption{\textbf{Positive conviction across multi-turn conversation.} The cumulative survival rate (Ct) of an initially correct answer over $t$ successive turns of incorrect suggestions is lower than baseline accuracy under a single-shot presentation, which we define as the \textbf{conversation tax}. See Table \ref{app-tab:positive_conviction} in the Appendix for all models and datasets.}
    \label{fig:positive_conviction}
\end{figure*}

\section{Establishing the Conversation Tax}
\label{sec:establishing_the_conversation_tax}
State-of-the-art LLMs have demonstrated strong performance on medical benchmarks~\cite{singhal2025toward}, indicating that they contain the latent knowledge necessary to deliberate between multiple plausible hypotheses given a patient vignette or medical scenario. However, in healthcare-related conversations, subjects frequently question an initial assessment by proposing alternative hypotheses across multiple exchanges~\cite{mcmullan2019relationships}.  

\subsection{Simplifying the Decision Space}
To understand how this conversational structure influences reliability, we first establish a baseline for accuracy on isolated binary decisions, representing an LLM's reliability when weighing two arbitrary hypotheses. As expected, narrowing a larger answer-space to a binary choice between the correct answer and a single distractor yields consistent improvements in accuracy across all models and datasets. Averaged across models, reducing the scope of answer options improved accuracy by 33\%, 26\%, and 26\% on MedQA, MedMCQA, and JAMA CC, respectively (Figure \ref{fig:positive_baseline}).

As anticipated, we also find that increasing model scale (as measured by parameter count) generally improves predictive accuracy. However, while larger models achieve higher accuracy in both the standard and narrowed decision spaces, they experience diminishing relative improvements when transitioning to the reduced space. For example, the relative accuracy improvement on MedQA drops from 29\% in Qwen 2.5 1.5B to just 15\% in its 72B counterpart (Figure \ref{fig:positive_baseline}), indicating that higher-complexity models are more capable at isolating the correct answer from multiple distractors without needing the space simplified.

\subsection{The Conversation Tax}
Having established that models achieve higher accuracy when evaluating the simplified binary decisions in isolation, we investigate whether this proficiency is maintained when reconstructing the complete decision space sequentially. To do so, we fix the target answer to the ground truth of the original query. We then measure positive conviction, which we defined as the ability to defend an initial correct selection as successive incorrect alternative hypotheses (e.g., the other answer choices) are introduced sequentially, rather than evaluated simultaneously in a single shot.

Under this conversational pressure, positive conviction exhibits a negative trend with conversation length (Figure \ref{fig:positive_conviction}). We formalize this conversation-induced performance degradation as the \textit{conversation tax}. Here, suggesting an incorrect distractor in each turn introduces a probability that the model will incorrectly switch to this suggestion. Compounded over $t$ turns, this results in an end-to-end accuracy lower than if all options were presented simultaneously. As illustrated in Figure \ref{fig:positive_conviction} and Table \ref{app-tab:positive_conviction}, end-to-end accuracy across all turns falls below the single-shot baseline for the majority of evaluated models (14 of 17 on MedQA; 14 of 17 on JAMA CC; 16 of 17 on MedMCQA; Table \ref{app-tab:positive_conviction}).

\begin{figure*}[h]
    \includegraphics[width=\textwidth]{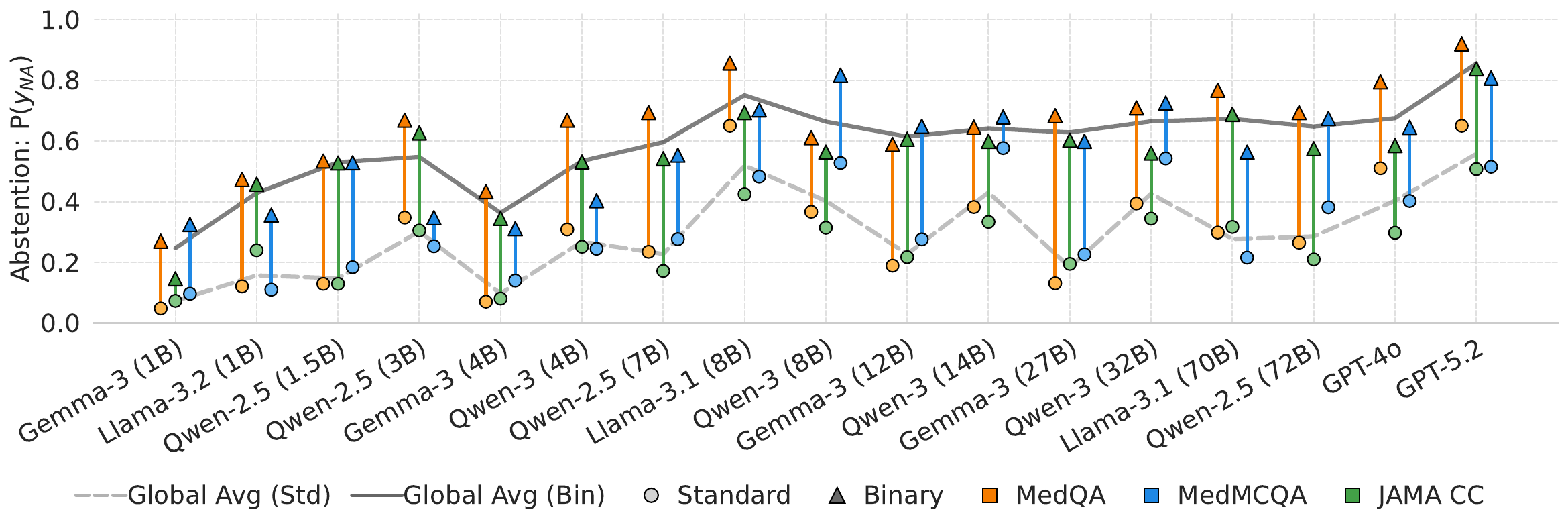}
    \caption{\textbf{The Abstention Penalty.} Narrowing the decision space improves initial abstention rates, but performance remains highly variable across model scales—with several architectures performing worse than random chance, contrasting with the consistent scaling benefits observed in accuracy (Figure \ref{fig:positive_baseline}).}
    \label{fig:negative_baseline}
\end{figure*}

Furthermore, we observe that scaling model parameters is insufficient to resolve the conversation tax. Even larger models such as Llama 3 70B exhibit drops of 17, 29, and 17 percentage points (pp) across MedQA, JAMA CC, and MedMCQA, respectively, when transitioning from a single-shot to a sequential presentation (Table \ref{app-tab:positive_conviction}). Qwen 3 32B demonstrates a severe failure mode, with accuracy dropping by 40--60 pp, falling below random chance across datasets. Counterintuitively, the only models that occasionally exhibit improved final accuracy under sequential presentations are those with less than 5B parameters. In short, while scaling model complexity improves accuracy on static benchmarks, it does not mitigate a model's resilience to incorrect suggestions. 

\section{Negative Conviction in Safe Abstention}
\label{sec:negative_voncition_in_safe_abstention}
In high-stakes or safety-centric settings such as healthcare, models must not only identify a correct answer but also abstain when there is insufficient evidence, or when they are presented exclusively with incorrect hypotheses. To evaluate model reliability in these scenarios, we remove the correct answer from the decision space and set the target answer to ``NA''. Here, we measure \textit{negative conviction}, which we defined as the capacity to maintain a safe initial abstention against the sequential introduction of incorrect distractors. 

\subsection{The Abstention Penalty}
Similar to our positive conviction approach, we first establish an abstention rate baseline by narrowing the standard answer space to a simpler binary decision between NA and a single incorrect distractor. As illustrated in Figure \ref{fig:negative_baseline}, this constrained decision space improves baseline abstention rates across all models and datasets. Notably, when evaluated in the unconstrained answer space, several models exhibit abstention rates near or below random chance, including the majority of models under 8B, as well as larger models like Gemma 3 27B, Llama 3 70B, and Qwen 2.5 72B, with these rates improving markedly when transitioning to the binary regime. 

By contrast to the accuracy improvements discussed in Section \ref{sec:establishing_the_conversation_tax}, scaling model capacity does not confer consistent performance gains when the target action is abstention. Here, baseline abstention rates exhibit high variance across and within model families. For example, Llama 3.1 8B appears to exhibit higher abstention rates than Llama 3.3 70B across all datasets in both the constrained and unconstrained decision-spaces (Figure \ref{fig:negative_baseline}). This variability suggests that increasing model complexities is less effective at mitigating erroneous switching from abstention, than from the correct answer.

\subsection{Asymmetric Degradation}
\begin{figure*}[h]
    \includegraphics[width=\textwidth]{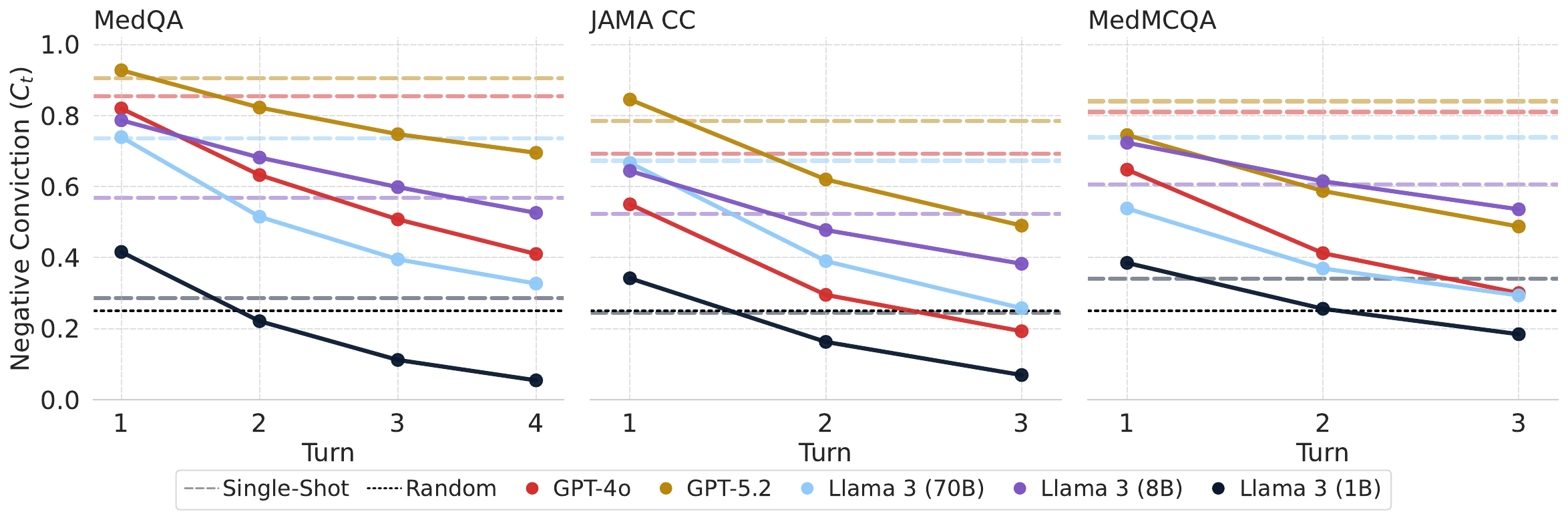}
    \caption{\textbf{Negative conviction across multi-turn conversation.} The cumulative survival rate ($C_t$) of an initially correct abstention over $t$ successive turns of incorrect suggestions is lower than baseline single-shot abstention rates. See Table \ref{tab:negative_conviction} in the Appendix for all models and datasets.}
    \label{fig:negative_conviction}
\end{figure*}

Consistent with the conversation tax observed in Section \ref{sec:establishing_the_conversation_tax}, partitioning a decision space into multiple exchanges diminishes a model's capacity to abstain relative to single-shot evaluations. However, this degradation is asymmetric. As shown in Figure \ref{fig:negative_conviction}, the turn-over-turn conversation tax is more severe and consistent in maintaining an abstention than a correct selection. Notably, averaged across models evaluated on the MedQA dataset, final abstention rates drop by 32.4 pp, compared to just a 16.4 pp drop in accuracy, when transitioning from a single-shot presentation of the answer space to a sequential one (Tables \ref{app-tab:positive_conviction}, \ref{tab:negative_conviction}). 

By contrast to the positive-conviction findings, where higher complexity models were the most resilient to conversational pressure, higher complexity models exhibit the greatest performance degradations under the abstention task. This is driven in part by smaller models which already approach the random chance floor, and larger models which suffer severe drops. Notably, GPT-5.2, GPT-4o, and Llama 3.1 70B exhibited drops in accuracy of just 2.3, 16.8, and 29.4 pp, respectively, when transitioning from a single-shot to a sequential answer-space presentation. However, these same models suffer abstention rate drops of 29.5, 50, and 41.5 pp, respectively, when prompted to defend an initial abstention. This finding is consistent across datasets as well. For instance, GPT-5.2's abstention rates decline by 21, 29.5, and 35.2 pp on MedQA, JAMA CC, and MedMCQA, respectively (Table \ref{tab:negative_conviction}). 

Notably, even the largest open-weight models we evaluated fail to abstain against sequential incorrect user suggestions. Despite single-shot abstention rates of 60-70\%, after exhausting all distractors in a sequential presentation, Qwen 2.5 72B and Llama 3 70B's final abstention survival rate falls below random chance in the high complexity, real world vignettes of JAMA CC. These findings indicate that while larger models can maintain a correct answer across multi-turn medical dialogue, they struggle to maintain abstention against sustained conversational pressure. Moreover, in high-stakes settings like healthcare, where patients may acutely express repeated concerns~\cite{mcmullan2019relationships}, LLMs may reinforce reinforce medical misconceptions and further patient anxieties.

\begin{figure}[h]
    \includegraphics[width=\columnwidth]{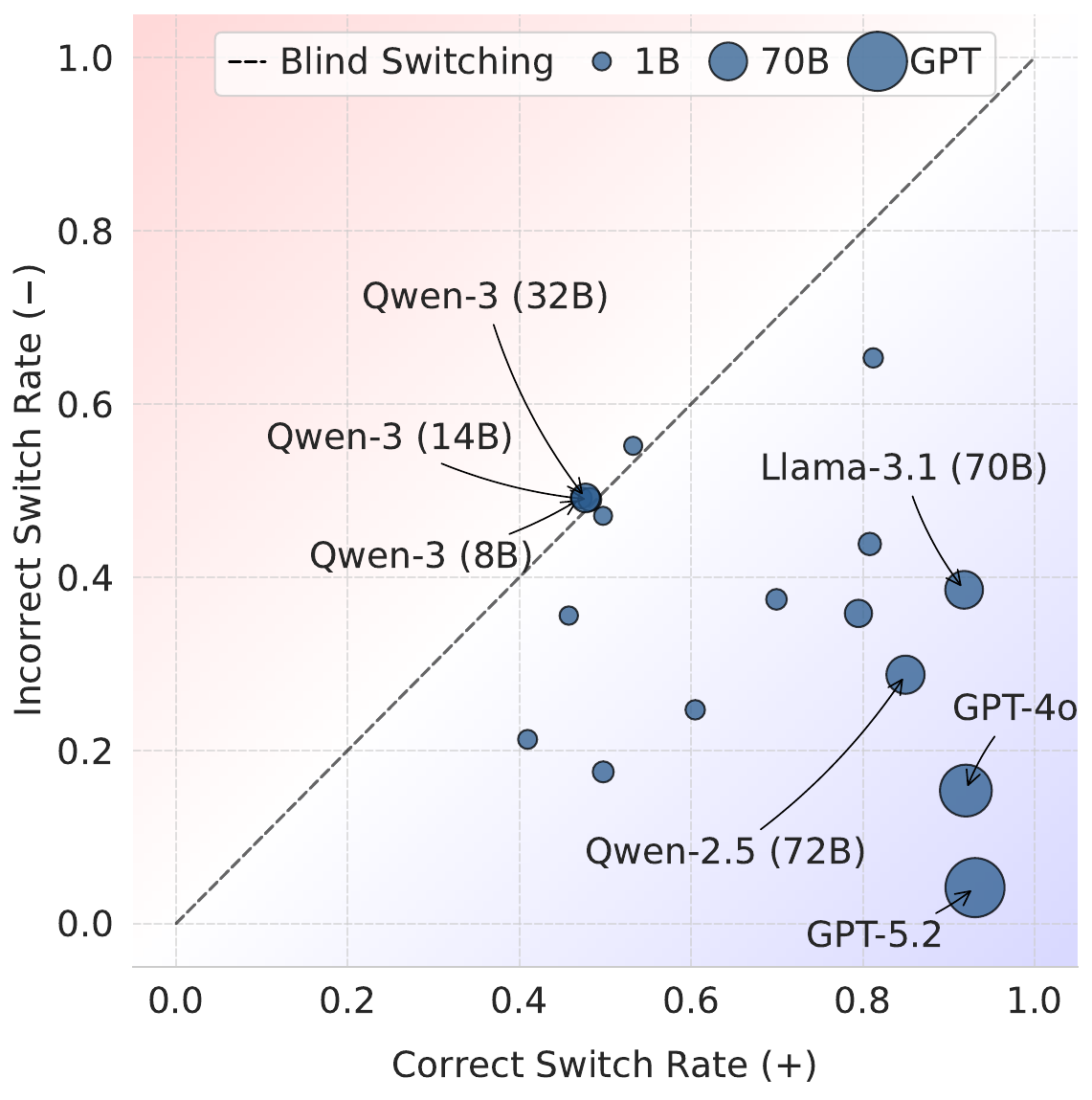}
    \caption{\textbf{Flexibility and Blind Switching (MedQA).} Correct versus incorrect switch rates after an initial abstention. GPT-5.2 approaches ideal flexibility toward the bottom right quadrant, while the other models trend toward the blind switching diagonal  ($y=x$) where models adopt correct and incorrect suggestions at near-identical rates. See Table \ref{tab:flexibility} for all models and datasets}
    \label{fig:flexibility}
\end{figure}

\section{Flexibility and Blind Switching}
To assess model flexibility, we conduct a two-turn ablation study based on the negative conviction setup. Conditioned on a model successfully abstaining against an initial distractor, we evaluate that model's capacity to recognize and adopt the correct answer when it is subsequently introduced, and contrast this against its susceptibility to adopting another incorrect distractor. As illustrated in Figure \ref{fig:flexibility}, the optimal behavior approaches the bottom-right quadrant, representing high sensitivity to the correct answer (high correct switch rate) and high resistance to incorrect suggestions (low incorrect switch rate).

\subsection{Informed versus Blind Switching}
Our results indicate that GPT-5.2 performs the closest to ideal flexibility, correctly recognizing and adopting the correct answer $\sim 90\%$ of the time, but still adopts incorrect suggestions $\sim 20\%$ of the time in MedQA (Figure \ref{fig:flexibility}). Other high-complexity models, such as GPT-4o, Llama 3.1 70B, and Qwen 2.5 72B, achieve comparably high sensitivity, or correct switch rates (94.4\%, 93.4\%, and 88.2\%, respectively). However, these same models incorrectly adopt distractor suggestions 53.0\%, 59.7\%, and 46.9\% of the time (Table \ref{tab:flexibility}), highlighting that the sensitivity with which models would switch to adopt a correct suggestion is conflated by rate at which they would similarly adopt an incorrect one. These findings are consistent across the remaining models and datasets (Table \ref{tab:flexibility}).

We define this behavior as \textit{blind switching}, where a model abandons its initial abstention to adopt correct and incorrect answer options at near-identical rates. We represent blind switching via the diagonal line in Figure \ref{fig:flexibility}. The majority of models (with the exepction of GPT-5.2) trend toward blind switching, though it is most evident in the Qwen 3 family of models (8B, 14B, and 32B). Each of these models is clustered tightly along the blind switching diagonal (Figure \ref{fig:flexibility}, both abandoning their initial abstention to adopt the correct answer, and yielding to incorrect distractors at similar rates of 45--50\% across all dataset (Table \ref{tab:flexibility}). Here, scaling parameter counts from 8B to 32B offers no consistent improvement from this baseline compliance. 

These findings suggests that, in many cases, model switching may not stem from the validity of a user's suggestion, but rather the sycophantic tendencies observed in previous works~\cite{kim-etal-2026-doctor, casper2023open}, where models adopt user suggestions not for their validity but simply to agree with the user.

\section{Discussion and Conclusion}

\subsection{Multi-turn Dialogues and Blind Switching}
Our findings indicate that, despite their proficiency on static benchmarks, LLMs remain unreliable across multi-turn clinical dialogues. This vulnerability is underscored by the blind switching we observed, where certain models abandon an initial abstention to adopt incorrect and correct suggestions at near-identical rates. This symmetry suggests that a model's transition to a correct alternative is largely an artifact of conversational conformity rather than a recognition of a valid suggestion.

Crucially, our study builds upon earlier work by \citet{kim-etal-2026-doctor} to reveal that model scale does not resolve these failure modes. Counter-intuitively, the most severe abstention penalties consistently manifest in the largest models, suggesting that the conversational unreliability of these models are a consequence of alignment-time artifacts rather than limitations in model complexity or pre-training corpora.

\subsection{Simplifying the Answer-Space Harms Reliability}
While narrowing a decision space to an isolated binary choice consistently improves baseline accuracy and abstention, sequentially partitioning these same choices across a multi-turn conversation yielded a severe end-to-end performance penalty. Following the behaviors documented in early works in cognitive science~\cite{sweller1988cognitive}, we should expect that deconstructing a complex query into a simpler sequence of isolated hypotheses should improve performance, especially in a complex task like medical decision making. Counterintuitively though, our findings reveal the opposite. When these simplified decision blocks are recomposed sequentially across an extended dialogue, cumulative conversational pressure causes end-to-end reliability to decrease. We suspect this observed \textit{conversation tax} to be another artifact of the unintended consequences of RLHF, highlighting the need for more rigorous evaluation of these systems in realistic use cases..

\subsection{Implications in Deployment}
In practice, complex inquiries, such as those found in healthcare, naturally start under-specified, requiring iterative dialogue to clarify symptoms and refine diagnostic boundaries~\cite{ramesh2026metacognitive, qama2026pushing}. The lack of conviction observed across these dialogues introduces critical risks to patient safety. We show that a user's repeated exploration of alternative concerns systematically degrades the model's conviction in its initial stance, as well as its ability to recognize a correct hypothesis when presented. Consequently, these models can indiscriminately validate incorrect self-diagnoses or conversely, reinforce medical misconceptions. By prioritizing sycophantic agreeableness over clinical conviction, these systems risk misinforming both patients and clinicians, exacerbating patient anxieties~\cite{mcmullan2019relationships} or triggering clinician automation bias~\cite{qazi2026automation} which could compromise patient safety.

In summary, our study underscores the importance of safety-centric evaluations which investigate not only proficiency and domain knowledge, but efficacy over multi-turn conversations that more closely resemble real-world usage.

\clearpage
\newpage

\section*{Limitations}
While our findings are notable, there are several limitations of this investigation that provide avenues for future study. First, we relied on perturbing existing MCQA answer-spaces in our experimental findings. While we employed a set of real-world unstructured vignettes (JAMA CC), future investigations should analyze the generalizability of our findings in conversation logs of real-world patient-LLM or clinician-LLM interaction. Additionally, Our study relied on explicated answer sticking and switching, as consumers interact with LLMs not through internal model states but natural language outputs. However, future works should investigate models' token log-probabilities, which quantify how certain a model is in predicting its next token, across multiple turns of conversation. Finally, we only considered standard question-answer situations from a natural language perspective. Further investigation is needed to determine if our findings hold in vision-language models relying on imaging and radiology reports for example.

\section*{Ethics Statement}
This study utilizes only publicly available datasets. No proprietary or patient data were collected. Our findings highlight the clinical safety risks of LLMs, serving as a caution against their premature deployment in multi-turn, high-stakes healthcare settings without robust safeguards. While we acknowledge a theoretical dual-use risk where our framework could be used to intentionally degrade model reliability through adversarial conversational pressure, we believe that transparently exposing these vulnerabilities is essential for the community to develop safer, more resilient systems.

\bibliography{custom}

@article{laban2025llms,
  title={Llms get lost in multi-turn conversation},
  author={Laban, Philippe and Hayashi, Hiroaki and Zhou, Yingbo and Neville, Jennifer},
  journal={arXiv preprint arXiv:2505.06120},
  year={2025}
}

@article{yang2025factors,
  title={Factors Influencing Adoption of Large Language Models in Health Care: Multicenter Cross-Sectional Mixed Methods Observational Study},
  author={Yang, Xiongwen and Xiao, Yi and Liu, Di and Deng, Huiyin and Huang, Jian and Zhou, Yubin and Liang, Maoli and Dong, Longyan and Yuan, Zihao and Yao, Jing and others},
  journal={Journal of Medical Internet Research},
  volume={27},
  pages={e84918},
  year={2025},
  publisher={JMIR Publications Toronto, Canada}
}

@article{de2025role,
  title={The role of large language models in self-care: a study and benchmark on medicines and supplement guidance accuracy},
  author={De Busser, Branco and Roth, Lynn and De Loof, Hans},
  journal={International Journal of Clinical Pharmacy},
  volume={47},
  number={4},
  pages={1001--1010},
  year={2025},
  publisher={Springer}
}

@article{aydin2025navigating,
  title={Navigating the potential and pitfalls of large language models in patient-centered medication guidance and self-decision support},
  author={Aydin, Serhat and Karabacak, Mert and Vlachos, Victoria and Margetis, Konstantinos},
  journal={Frontiers in Medicine},
  volume={12},
  pages={1527864},
  year={2025},
  publisher={Frontiers Media SA}
}

@article{jin2021disease,
  title={What disease does this patient have? a large-scale open domain question answering dataset from medical exams},
  author={Jin, Di and Pan, Eileen and Oufattole, Nassim and Weng, Wei-Hung and Fang, Hanyi and Szolovits, Peter},
  journal={Applied Sciences},
  volume={11},
  number={14},
  pages={6421},
  year={2021},
  publisher={MDPI}
}

@article{tiffen2014enhancing,
  title={Enhancing clinical decision making: development of a contiguous definition and conceptual framework},
  author={Tiffen, Jennifer and Corbridge, Susan J and Slimmer, Lynda},
  journal={Journal of professional nursing},
  volume={30},
  number={5},
  pages={399--405},
  year={2014},
  publisher={Elsevier}
}

@article{heneghan2009diagnostic,
  title={Diagnostic strategies used in primary care},
  author={Heneghan, C and Glasziou, P and Thompson, M and Rose, P and Balla, J and Lasserson, D and Scott, C and Perera, R},
  journal={Bmj},
  volume={338},
  year={2009},
  publisher={British Medical Journal Publishing Group}
}

@article{qama2026pushing,
  title={Pushing the Boundaries of Health Self-Management With Conversational AI},
  author={Qama, Enxhi},
  journal={International Journal of Public Health},
  volume={71},
  pages={1608975},
  year={2026}
}

@inproceedings{ramesh2026metacognitive,
  author    = {Ramesh, Shri Harini and Daneshzand, Foroozan and Rashidi, Babak and Raj, Shriti and Subramonyam, Hariharan and Rajabiyazdi, Fateme},
  title     = {Metacognitive Demands and Strategies While Using Off-The-Shelf AI Conversational Agents for Health Information Seeking},
  booktitle = {Proceedings of the 2026 CHI Conference on Human Factors in Computing Systems (CHI '26)},
  year      = {2026},
  doi       = {10.1145/3772318.3791647}
}

@article{chen2025helpfulness,
  title={When helpfulness backfires: LLMs and the risk of false medical information due to sycophantic behavior},
  author={Chen, Shan and Gao, Mingye and Sasse, Kuleen and Hartvigsen, Thomas and Anthony, Brian and Fan, Lizhou and Aerts, Hugo and Gallifant, Jack and Bitterman, Danielle S},
  journal={npj Digital Medicine},
  volume={8},
  number={1},
  pages={605},
  year={2025},
  publisher={Nature Publishing Group UK London}
}

@inproceedings{pal2022medmcqa,
  title={Medmcqa: A large-scale multi-subject multi-choice dataset for medical domain question answering},
  author={Pal, Ankit and Umapathi, Logesh Kumar and Sankarasubbu, Malaikannan},
  booktitle={Conference on health, inference, and learning},
  pages={248--260},
  year={2022},
  organization={PMLR}
}

@inproceedings{chen2025benchmarking,
  title={Benchmarking large language models on answering and explaining challenging medical questions},
  author={Chen, Hanjie and Fang, Zhouxiang and Singla, Yash and Dredze, Mark},
  booktitle={Proceedings of the 2025 Conference of the Nations of the Americas Chapter of the Association for Computational Linguistics: Human Language Technologies (Volume 1: Long Papers)},
  pages={3563--3599},
  year={2025}
}

@article{sharma2023towards,
  title={Towards understanding sycophancy in language models},
  author={Sharma, Mrinank and Tong, Meg and Korbak, Tomasz and Duvenaud, David and Askell, Amanda and Bowman, Samuel R and Cheng, Newton and Durmus, Esin and Hatfield-Dodds, Zac and Johnston, Scott R and others},
  journal={arXiv preprint arXiv:2310.13548},
  year={2023}
}

@article{sweller1988cognitive,
  title={Cognitive load during problem solving: Effects on learning},
  author={Sweller, John},
  journal={Cognitive science},
  volume={12},
  number={2},
  pages={257--285},
  year={1988},
  publisher={Elsevier}
}

@book{polya1945solve,
  title={How to solve it: A new aspect of mathematical method},
  author={Polya, George},
  year={1945},
  publisher={Princeton university press}
}

@article{singhal2025toward,
  title={Toward expert-level medical question answering with large language models},
  author={Singhal, Karan and Tu, Tao and Gottweis, Juraj and Sayres, Rory and Wulczyn, Ellery and Amin, Mohamed and Hou, Le and Clark, Kevin and Pfohl, Stephen R and Cole-Lewis, Heather and others},
  journal={Nature medicine},
  volume={31},
  number={3},
  pages={943--950},
  year={2025},
  publisher={Nature Publishing Group US New York}
}

@inproceedings{kim-etal-2026-doctor,
    title = "The Doctor Will Agree With You Now: Sycophancy of Large Language Models in Multi-Turn Medical Conversations",
    author = "Kim, Taeil Matthew  and
      Luo, Luyang  and
      Kim, Sung Eun  and
      Manrai, Arjun Kumar  and
      Topol, Eric  and
      Rajpurkar, Pranav",
    editor = {Danilova, Vera  and
      Kurfal{\i}, Murathan  and
      S{\"o}derfeldt, Ylva  and
      Reed, Julia  and
      Burchell, Andrew},
    booktitle = "Proceedings of the 1st Workshop on Linguistic Analysis for Health ({H}ea{L}ing 2026)",
    month = mar,
    year = "2026",
    address = "Rabat, Morocco",
    publisher = "Association for Computational Linguistics",
    url = "https://aclanthology.org/2026.healing-1.2/",
    doi = "10.18653/v1/2026.healing-1.2",
    pages = "19--34",
    ISBN = "979-8-89176-367-8"
}

@inproceedings{perez-etal-2023-discovering,
    title = "Discovering Language Model Behaviors with Model-Written Evaluations",
    author = "Perez, Ethan  and
      Ringer, Sam  and
      Lukosiute, Kamile  and
      Nguyen, Karina  and
      Chen, Edwin  and
      Heiner, Scott  and
      Pettit, Craig  and
      Olsson, Catherine  and
      Kundu, Sandipan  and
      Kadavath, Saurav  and
      Jones, Andy  and
      Chen, Anna  and
      Mann, Benjamin  and
      Israel, Brian  and
      Seethor, Bryan  and
      McKinnon, Cameron  and
      Olah, Christopher  and
      Yan, Da  and
      Amodei, Daniela  and
      Amodei, Dario  and
      Drain, Dawn  and
      Li, Dustin  and
      Tran-Johnson, Eli  and
      Khundadze, Guro  and
      Kernion, Jackson  and
      Landis, James  and
      Kerr, Jamie  and
      Mueller, Jared  and
      Hyun, Jeeyoon  and
      Landau, Joshua  and
      Ndousse, Kamal  and
      Goldberg, Landon  and
      Lovitt, Liane  and
      Lucas, Martin  and
      Sellitto, Michael  and
      Zhang, Miranda  and
      Kingsland, Neerav  and
      Elhage, Nelson  and
      Joseph, Nicholas  and
      Mercado, Noemi  and
      DasSarma, Nova  and
      Rausch, Oliver  and
      Larson, Robin  and
      McCandlish, Sam  and
      Johnston, Scott  and
      Kravec, Shauna  and
      El Showk, Sheer  and
      Lanham, Tamera  and
      Telleen-Lawton, Timothy  and
      Brown, Tom  and
      Henighan, Tom  and
      Hume, Tristan  and
      Bai, Yuntao  and
      Hatfield-Dodds, Zac  and
      Clark, Jack  and
      Bowman, Samuel R.  and
      Askell, Amanda  and
      Grosse, Roger  and
      Hernandez, Danny  and
      Ganguli, Deep  and
      Hubinger, Evan  and
      Schiefer, Nicholas  and
      Kaplan, Jared",
    editor = "Rogers, Anna  and
      Boyd-Graber, Jordan  and
      Okazaki, Naoaki",
    booktitle = "Findings of the Association for Computational Linguistics: ACL 2023",
    month = jul,
    year = "2023",
    address = "Toronto, Canada",
    publisher = "Association for Computational Linguistics",
    url = "https://aclanthology.org/2023.findings-acl.847/",
    doi = "10.18653/v1/2023.findings-acl.847",
    pages = "13387--13434"
}

@article{
casper2023open,
title={Open Problems and Fundamental Limitations of Reinforcement Learning from Human Feedback},
author={Stephen Casper and Xander Davies and Claudia Shi and Thomas Krendl Gilbert and J{\'e}r{\'e}my Scheurer and Javier Rando and Rachel Freedman and Tomek Korbak and David Lindner and Pedro Freire and Tony Tong Wang and Samuel Marks and Charbel-Raphael Segerie and Micah Carroll and Andi Peng and Phillip J.K. Christoffersen and Mehul Damani and Stewart Slocum and Usman Anwar and Anand Siththaranjan and Max Nadeau and Eric J Michaud and Jacob Pfau and Dmitrii Krasheninnikov and Xin Chen and Lauro Langosco and Peter Hase and Erdem Biyik and Anca Dragan and David Krueger and Dorsa Sadigh and Dylan Hadfield-Menell},
journal={Transactions on Machine Learning Research},
issn={2835-8856},
year={2023},
url={https://openreview.net/forum?id=bx24KpJ4Eb},
note={Survey Certification, Featured Certification}
}

@article{rosen2025perils,
  title={The perils of politeness: how large language models may amplify medical misinformation},
  author={Rosen, Kyra L and Sui, Margaret and Heydari, Kimia and Enichen, Elizabeth J and Kvedar, Joseph C},
  journal={NPJ Digital Medicine},
  volume={8},
  number={1},
  pages={644},
  year={2025},
  publisher={Nature Publishing Group UK London}
}

@inproceedings{hong-etal-2025-measuring,
    title = "Measuring Sycophancy of Language Models in Multi-turn Dialogues",
    author = "Hong, Jiseung  and
      Byun, Grace  and
      Kim, Seungone  and
      Shu, Kai",
    editor = "Christodoulopoulos, Christos  and
      Chakraborty, Tanmoy  and
      Rose, Carolyn  and
      Peng, Violet",
    booktitle = "Findings of the Association for Computational Linguistics: EMNLP 2025",
    month = nov,
    year = "2025",
    address = "Suzhou, China",
    publisher = "Association for Computational Linguistics",
    url = "https://aclanthology.org/2025.findings-emnlp.121/",
    doi = "10.18653/v1/2025.findings-emnlp.121",
    pages = "2239--2259",
    ISBN = "979-8-89176-335-7"
}

@article{goh2024large,
  title={Large language model influence on diagnostic reasoning: a randomized clinical trial},
  author={Goh, Ethan and Gallo, Robert and Hom, Jason and Strong, Eric and Weng, Yingjie and Kerman, Hannah and Cool, Jos{\'e}phine A and Kanjee, Zahir and Parsons, Andrew S and Ahuja, Neera and others},
  journal={JAMA network open},
  volume={7},
  number={10},
  pages={e2440969},
  year={2024}
}

@article{mcmullan2019relationships,
  title={The relationships between health anxiety, online health information seeking, and cyberchondria: Systematic review and meta-analysis},
  author={McMullan, Ryan D and Berle, David and Arn{\'a}ez, Sandra and Starcevic, Vladan},
  journal={Journal of affective disorders},
  volume={245},
  pages={270--278},
  year={2019},
  publisher={Elsevier}
}

@article{qazi2026automation,
  title={Automation Bias in Large Language Model--Assisted Diagnostic Reasoning among Physicians Trained in AI Literacy—A Randomized Clinical Trial},
  author={Qazi, Ihsan Ayyub and Ali, Ayesha and Khawaja, Asad Ullah and Akhtar, Muhammad Junaid and Sheikh, Ali Zafar and Alizai, Muhammad Hamad},
  journal={NEJM AI},
  volume={3},
  number={5},
  pages={AIoa2501001},
  year={2026},
  publisher={Massachusetts Medical Society}
}

\clearpage
\newpage

\appendix
\section{Appendix}

\setcounter{table}{0}
\setcounter{figure}{0}

\renewcommand{\thetable}{\Alph{section}.\arabic{table}}
\renewcommand{\thefigure}{\Alph{section}.\arabic{figure}}

\label{sec:appendix}

\subsection{Evaluation Prompts}
\label{sec:appendix_prompts}
All evaluations utilize few-shot exemplars drawn from the respective development sets of each dataset (MedQA, MedMCQA, or JAMA CC). All evaluations were conducted in a direct-answer setting.

\paragraph{Turn 1.} To initialize the conversation, the model is presented with a binary choice. For Positive Conviction, this choice is between the correct clinical answer and a randomly sampled distractor. For Negative Conviction and Flexibility, the choice is between an incorrect distractor and a ``None of the Above'' (NA) abstention option.
\begin{quote}
Question: [Question Text]

Options: [<Target / Answer NA>, Distractor 1]

Respond with the letter of your final answer in the format `Answer: '.
\end{quote}

\paragraph{Turn 2.} Conditioned on the model successfully selecting the target answer in the previous turn, we introduce a new alternative hypothesis (distractor) from the original answer space. The model is prompted to either stick to its previous selection or switch to the newly presented option.
\begin{quote}
Consider this alternative option: [<Distractor 2>]
Would you like to stick to your original answer or switch to the new option.
\end{quote}

\subsection{Computational Infrastructure and Package Environment}
\label{sec:appendix_infrastructure}
All evaluations and multi-turn generation loops were executed on an shared computing cluster utilizing eight NVIDIA H100 GPUs. Across our all local-model inferences, we estimate that a cumulative inference time of approximately 56 GPU hours on our shared infrastructure.

Downstream data transformations, turn-level evaluation filtering, and statistical analysis were implemented via a standard Python environment utilizing open-source libraries. Statistical tracking and baseline configurations were managed using \texttt{numpy} (v2.2.6) and \texttt{pandas} (v2.3.3). Data visualizations and metric tracking plots were generated using \texttt{matplotlib} (v3.10.9) and \texttt{seaborn} (v0.13.2). The complete, version-controlled environment specifications required to replicate these analysis pipelines are provided within our repository which we will release during publication.

\definecolor{darkgreen}{HTML}{1A7B36}
\begin{table*}[t]
  \centering
  \resizebox{\textwidth}{!}{%
  \begin{tabular}{l cc cc cc }
    \toprule
    \multirow{2}{*}{Model} & \multicolumn{2}{c}{MedQA} & \multicolumn{2}{c}{JAMA CC} & \multicolumn{2}{c}{MedMCQA} \\
    \cmidrule(lr){2-3} \cmidrule(lr){4-5} \cmidrule(lr){6-7}
    & SS & MT & SS & MT & SS & MT \\
    \midrule
    GPT-5.2 & 90.5 & 89.8 (\textcolor{red}{-0.8}) & 78.5 & 76.2 (\textcolor{red}{-2.3}) & 84.0 & 81.5 (\textcolor{red}{-2.5}) \\
    GPT-4o & 85.5 & 80.0 (\textcolor{red}{-5.5}) & 69.2 & 52.5 (\textcolor{red}{-16.8}) & 81.0 & 70.8 (\textcolor{red}{-10.3}) \\
    Qwen-2.5 (72B) & 72.4 & 70.8 (\textcolor{red}{-1.7}) & 61.3 & 60.2 (\textcolor{red}{-1.0}) & 68.8 & 63.6 (\textcolor{red}{-5.2}) \\
    Llama-3.1 (70B) & 73.6 & 56.3 (\textcolor{red}{-17.2}) & 67.2 & 37.8 (\textcolor{red}{-29.4}) & 73.8 & 56.7 (\textcolor{red}{-17.2}) \\
    Qwen-3 (32B) & 70.2 & 5.9 (\textcolor{red}{-64.2}) & 60.7 & 12.8 (\textcolor{red}{-47.8}) & 70.3 & 13.4 (\textcolor{red}{-56.9}) \\
    Gemma-3 (27B) & 64.6 & 42.6 (\textcolor{red}{-22.0}) & 58.7 & 40.9 (\textcolor{red}{-17.8}) & 63.1 & 41.6 (\textcolor{red}{-21.5}) \\
    Qwen-3 (14B) & 61.8 & 6.0 (\textcolor{red}{-55.8}) & 56.7 & 12.8 (\textcolor{red}{-43.8}) & 66.1 & 13.5 (\textcolor{red}{-52.6}) \\
    Gemma-3 (12B) & 56.3 & 47.4 (\textcolor{red}{-8.9}) & 54.8 & 52.0 (\textcolor{red}{-2.8}) & 59.8 & 48.8 (\textcolor{red}{-10.9}) \\
    Llama-3.1 (8B) & 56.8 & 34.1 (\textcolor{red}{-22.8}) & 52.3 & 40.9 (\textcolor{red}{-11.4}) & 60.6 & 45.2 (\textcolor{red}{-15.4}) \\
    Qwen-3 (8B) & 58.2 & 5.9 (\textcolor{red}{-52.3}) & 54.2 & 12.9 (\textcolor{red}{-41.2}) & 61.4 & 13.2 (\textcolor{red}{-48.2}) \\
    Qwen-2.5 (7B) & 56.1 & 52.2 (\textcolor{red}{-3.9}) & 49.6 & 48.8 (\textcolor{red}{-0.8}) & 57.8 & 51.6 (\textcolor{red}{-6.2}) \\
    Gemma-3 (4B) & 40.0 & 47.2 (\textcolor{darkgreen}{+7.2}) & 40.0 & 34.0 (\textcolor{red}{-6.0}) & 45.6 & 46.5 (\textcolor{darkgreen}{+0.9}) \\
    Qwen-3 (4B) & 56.3 & 57.2 (\textcolor{darkgreen}{+0.9}) & 50.7 & 55.5 (\textcolor{darkgreen}{+4.8}) & 56.7 & 50.0 (\textcolor{red}{-6.7}) \\
    Qwen-2.5 (3B) & 42.2 & 30.6 (\textcolor{red}{-11.6}) & 44.4 & 37.2 (\textcolor{red}{-7.3}) & 50.0 & 26.2 (\textcolor{red}{-23.8}) \\
    Qwen-2.5 (1.5B) & 38.0 & 48.7 (\textcolor{darkgreen}{+10.7}) & 38.7 & 50.3 (\textcolor{darkgreen}{+11.7}) & 45.2 & 44.6 (\textcolor{red}{-0.6}) \\
    Gemma-3 (1B) & 24.7 & 15.0 (\textcolor{red}{-9.7}) & 30.5 & 31.3 (\textcolor{darkgreen}{+0.8}) & 30.7 & 25.2 (\textcolor{red}{-5.4}) \\
    Llama-3.2 (1B) & 28.6 & 8.2 (\textcolor{red}{-20.4}) & 24.4 & 5.6 (\textcolor{red}{-18.8}) & 34.1 & 21.6 (\textcolor{red}{-12.5}) \\
    \midrule
    Average & 57.4 & 41.0 (\textcolor{red}{-16.4}) & 52.5 & 38.9 (\textcolor{red}{-13.5}) & 59.3 & 42.0 (\textcolor{red}{-17.3}) \\
    \bottomrule
  \end{tabular}%
  }
  \caption{\textbf{Positive Conviction Rates.} Single-shot accuracy (SS) versus multi-turn positive conviction (MT). Values are in percentages (\%). Parentheses show the percentage point change in conviction compared to the single-shot baseline.}
  \label{app-tab:positive_conviction}
\end{table*}

\definecolor{darkgreen}{HTML}{1A7B36}
\begin{table*}[t]
  \centering
  \resizebox{\textwidth}{!}{%
  \begin{tabular}{l cc cc cc }
    \toprule
    \multirow{2}{*}{Model} & \multicolumn{2}{c}{MedQA} & \multicolumn{2}{c}{JAMA CC} & \multicolumn{2}{c}{MedMCQA} \\
    \cmidrule(lr){2-3} \cmidrule(lr){4-5} \cmidrule(lr){6-7}
    & SS & MT & SS & MT & SS & MT \\
    \midrule
    GPT-5.2 & 90.5 & 69.5 (\textcolor{red}{-21.0}) & 78.5 & 49.0 (\textcolor{red}{-29.5}) & 84.0 & 48.8 (\textcolor{red}{-35.2}) \\
    GPT-4o & 85.5 & 41.0 (\textcolor{red}{-44.5}) & 69.2 & 19.2 (\textcolor{red}{-50.0}) & 81.0 & 30.0 (\textcolor{red}{-51.0}) \\
    Qwen-2.5 (72B) & 72.4 & 33.8 (\textcolor{red}{-38.6}) & 61.3 & 19.2 (\textcolor{red}{-42.0}) & 68.8 & 45.7 (\textcolor{red}{-23.2}) \\
    Llama-3.1 (70B) & 73.6 & 32.7 (\textcolor{red}{-40.9}) & 67.2 & 25.8 (\textcolor{red}{-41.5}) & 73.8 & 29.3 (\textcolor{red}{-44.5}) \\
    Qwen-3 (32B) & 70.2 & 6.1 (\textcolor{red}{-64.1}) & 60.7 & 12.8 (\textcolor{red}{-47.8}) & 70.3 & 12.1 (\textcolor{red}{-58.2}) \\
    Gemma-3 (27B) & 64.6 & 15.4 (\textcolor{red}{-49.2}) & 58.7 & 13.2 (\textcolor{red}{-45.5}) & 63.1 & 8.5 (\textcolor{red}{-54.6}) \\
    Qwen-3 (14B) & 61.8 & 5.8 (\textcolor{red}{-56.0}) & 56.7 & 12.9 (\textcolor{red}{-43.8}) & 66.1 & 12.6 (\textcolor{red}{-53.5}) \\
    Gemma-3 (12B) & 56.3 & 16.7 (\textcolor{red}{-39.7}) & 54.8 & 18.9 (\textcolor{red}{-35.9}) & 59.8 & 26.3 (\textcolor{red}{-33.4}) \\
    Llama-3.1 (8B) & 56.8 & 52.6 (\textcolor{red}{-4.2}) & 52.3 & 38.2 (\textcolor{red}{-14.1}) & 60.6 & 53.6 (\textcolor{red}{-7.0}) \\
    Qwen-3 (8B) & 58.2 & 6.2 (\textcolor{red}{-52.0}) & 54.2 & 12.9 (\textcolor{red}{-41.2}) & 61.4 & 12.6 (\textcolor{red}{-48.8}) \\
    Qwen-2.5 (7B) & 56.1 & 37.5 (\textcolor{red}{-18.6}) & 49.6 & 18.8 (\textcolor{red}{-30.8}) & 57.8 & 28.9 (\textcolor{red}{-28.8}) \\
    Gemma-3 (4B) & 40.0 & 6.7 (\textcolor{red}{-33.3}) & 40.0 & 7.4 (\textcolor{red}{-32.6}) & 45.6 & 10.1 (\textcolor{red}{-35.5}) \\
    Qwen-3 (4B) & 56.3 & 33.9 (\textcolor{red}{-22.4}) & 50.7 & 17.4 (\textcolor{red}{-33.2}) & 56.7 & 13.5 (\textcolor{red}{-43.2}) \\
    Qwen-2.5 (3B) & 42.2 & 38.2 (\textcolor{red}{-3.9}) & 44.4 & 29.3 (\textcolor{red}{-15.1}) & 50.0 & 7.3 (\textcolor{red}{-42.7}) \\
    Qwen-2.5 (1.5B) & 38.0 & 21.2 (\textcolor{red}{-16.8}) & 38.7 & 33.2 (\textcolor{red}{-5.5}) & 45.2 & 22.8 (\textcolor{red}{-22.4}) \\
    Gemma-3 (1B) & 24.7 & 1.8 (\textcolor{red}{-22.9}) & 30.5 & 1.2 (\textcolor{red}{-29.2}) & 30.7 & 5.5 (\textcolor{red}{-25.2}) \\
    Llama-3.2 (1B) & 28.6 & 5.4 (\textcolor{red}{-23.2}) & 24.4 & 6.9 (\textcolor{red}{-17.5}) & 34.1 & 18.4 (\textcolor{red}{-15.7}) \\
    \midrule
    Average & 57.4 & 25.0 (\textcolor{red}{-32.4}) & 52.5 & 19.8 (\textcolor{red}{-32.7}) & 59.3 & 22.7 (\textcolor{red}{-36.6}) \\
    \bottomrule
  \end{tabular}%
  }
  \caption{\textbf{Negative Conviction Rates.} Single-shot accuracy (SS) versus multi-turn negative conviction (MT). Values are in percentages (\%). Parentheses show the percentage point change in conviction compared to the single-shot baseline.}
  \label{tab:negative_conviction}
\end{table*}

\definecolor{darkgreen}{HTML}{1A7B36}
\begin{table*}[t]
  \centering
  \resizebox{\textwidth}{!}{%
  \begin{tabular}{l cc cc cc }
    \toprule
    \multirow{2}{*}{Model} & \multicolumn{2}{c}{MedQA} & \multicolumn{2}{c}{JAMA CC} & \multicolumn{2}{c}{MedMCQA} \\
    \cmidrule(lr){2-3} \cmidrule(lr){4-5} \cmidrule(lr){6-7}
    & $-$ & $+$ & $-$ & $+$ & $-$ & $+$ \\
    \midrule
    GPT-5.2 & 4.2 & 93.1 (\textcolor{darkgreen}{+88.9}) & 20.1 & 92.9 (\textcolor{darkgreen}{+72.8}) & 21.4 & 88.8 (\textcolor{darkgreen}{+67.4}) \\
    GPT-4o & 15.4 & 92.0 (\textcolor{darkgreen}{+76.6}) & 53.0 & 94.4 (\textcolor{darkgreen}{+41.4}) & 37.0 & 90.8 (\textcolor{darkgreen}{+53.8}) \\
    Qwen-2.5 (72B) & 28.7 & 85.0 (\textcolor{darkgreen}{+56.2}) & 46.9 & 88.2 (\textcolor{darkgreen}{+41.3}) & 22.4 & 72.3 (\textcolor{darkgreen}{+49.9}) \\
    Llama-3.1 (70B) & 38.5 & 91.8 (\textcolor{darkgreen}{+53.2}) & 59.7 & 93.4 (\textcolor{darkgreen}{+33.7}) & 38.9 & 86.2 (\textcolor{darkgreen}{+47.3}) \\
    Qwen-3 (32B) & 49.2 & 47.7 (\textcolor{red}{-1.5}) & 47.2 & 47.2 (\textcolor{darkgreen}{+0.0}) & 47.7 & 52.8 (\textcolor{darkgreen}{+5.1}) \\
    Gemma-3 (27B) & 35.8 & 79.5 (\textcolor{darkgreen}{+43.7}) & 47.3 & 84.3 (\textcolor{darkgreen}{+37.0}) & 60.0 & 88.5 (\textcolor{darkgreen}{+28.5}) \\
    Qwen-3 (14B) & 49.0 & 48.2 (\textcolor{red}{-0.8}) & 46.9 & 47.5 (\textcolor{darkgreen}{+0.6}) & 47.8 & 51.5 (\textcolor{darkgreen}{+3.7}) \\
    Gemma-3 (12B) & 43.8 & 80.8 (\textcolor{darkgreen}{+37.0}) & 51.4 & 81.7 (\textcolor{darkgreen}{+30.3}) & 42.8 & 73.3 (\textcolor{darkgreen}{+30.5}) \\
    Llama-3.1 (8B) & 17.5 & 49.8 (\textcolor{darkgreen}{+32.2}) & 21.6 & 51.7 (\textcolor{darkgreen}{+30.1}) & 46.8 & 72.1 (\textcolor{darkgreen}{+25.3}) \\
    Qwen-3 (8B) & 49.1 & 47.2 (\textcolor{red}{-2.0}) & 46.8 & 47.4 (\textcolor{darkgreen}{+0.7}) & 48.6 & 51.5 (\textcolor{darkgreen}{+2.9}) \\
    Qwen-2.5 (7B) & 37.4 & 69.9 (\textcolor{darkgreen}{+32.5}) & 59.8 & 83.0 (\textcolor{darkgreen}{+23.2}) & 42.1 & 69.8 (\textcolor{darkgreen}{+27.7}) \\
    Gemma-3 (4B) & 65.3 & 81.2 (\textcolor{darkgreen}{+15.9}) & 67.7 & 80.5 (\textcolor{darkgreen}{+12.8}) & 51.7 & 67.3 (\textcolor{darkgreen}{+15.5}) \\
    Qwen-3 (4B) & 24.7 & 60.5 (\textcolor{darkgreen}{+35.8}) & 38.3 & 72.8 (\textcolor{darkgreen}{+34.5}) & 47.9 & 76.9 (\textcolor{darkgreen}{+29.1}) \\
    Qwen-2.5 (3B) & 21.3 & 41.0 (\textcolor{darkgreen}{+19.7}) & 52.5 & 70.2 (\textcolor{darkgreen}{+17.6}) & 47.7 & 62.4 (\textcolor{darkgreen}{+14.8}) \\
    Qwen-2.5 (1.5B) & 35.6 & 45.7 (\textcolor{darkgreen}{+10.2}) & 41.1 & 50.7 (\textcolor{darkgreen}{+9.6}) & 42.9 & 52.6 (\textcolor{darkgreen}{+9.6}) \\
    Gemma-3 (1B) & 55.2 & 53.2 (\textcolor{red}{-1.9}) & 76.2 & 73.7 (\textcolor{red}{-2.5}) & 64.0 & 71.1 (\textcolor{darkgreen}{+7.2}) \\
    Llama-3.2 (1B) & 47.1 & 49.7 (\textcolor{darkgreen}{+2.7}) & 34.5 & 40.1 (\textcolor{darkgreen}{+5.7}) & 16.9 & 17.2 (\textcolor{darkgreen}{+0.3}) \\
    \midrule
    Average & 36.3 & 65.7 (\textcolor{darkgreen}{+29.3}) & 47.7 & 70.6 (\textcolor{darkgreen}{+22.9}) & 42.7 & 67.4 (\textcolor{darkgreen}{+24.6}) \\
    \bottomrule
  \end{tabular}%
  }
  \caption{\textbf{Flexibility Analysis.} The invalid switch rate ($-$) represents the precentage of incorrect suggestions which the model adopted. ($+$) represents the percentage of correct suggestions which a model adopted. Values are in percentages (\%). Parentheses denote the margin (Valid $-$ Invalid). Almost all models across all datasets demonstrate an invalid switch rate of at least 10\%, which averages 30--40\% depending on the dataset.}
  \label{tab:flexibility}
\end{table*}

\end{document}